\documentclass[10pt,twocolumn,letterpaper]{article}

\usepackage{iccv}
\usepackage{times}
\usepackage{epsfig}
\usepackage{graphicx}
\usepackage{amsmath}
\usepackage{amssymb}

\usepackage{booktabs}
\usepackage{siunitx}
\usepackage{algorithm}
\usepackage{algpseudocode}
\usepackage{bbm}
\usepackage{subcaption}
\usepackage{multirow}
\usepackage{authblk}

\usepackage[pagebackref=true,breaklinks=true,colorlinks,bookmarks=false]{hyperref}

\iccvfinalcopy %
\def\iccvPaperID{905} %

\ificcvfinal\fi
\begin{document}
\title{\vspace{-1.0cm} Free-Form Image Inpainting with Gated Convolution\vspace{-0.5cm} }

\author[1]{Jiahui Yu}
\author[2]{Zhe Lin}
\author[2]{Jimei Yang}
\author[3]{Xiaohui Shen}
\author[2]{Xin Lu}
\author[1]{Thomas Huang}

\affil[1]{University of Illinois at Urbana-Champaign}
\affil[2]{Adobe Research}
\affil[3]{ByteDance AI Lab}
\renewcommand\Authands{\ \ \ \ \ }
\renewcommand{\Authsep}{\ \ \ \ \ }

\twocolumn[{%
\renewcommand\twocolumn[1][]{#1}%
\maketitle

\ificcvfinal\thispagestyle{empty}\fi

\begin{center}
\noindent
\vspace*{-1.2cm}
\includegraphics[width=\textwidth]{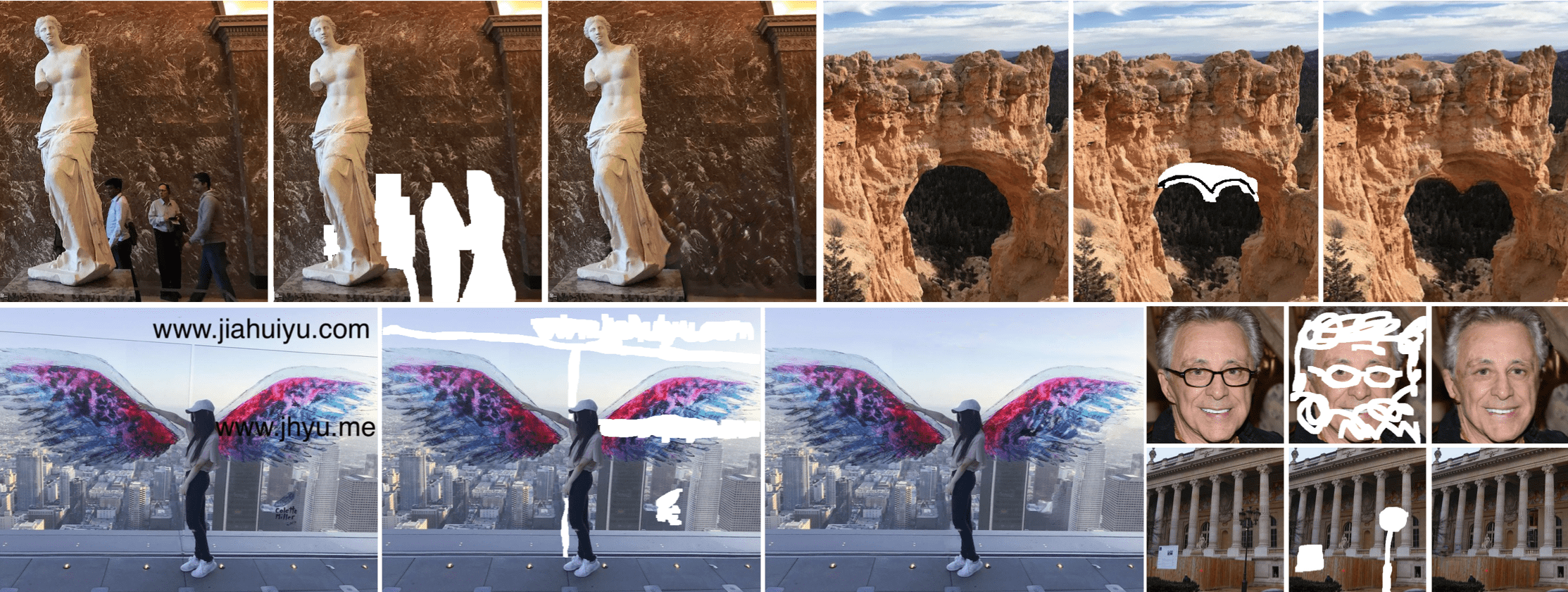}
\vspace*{-5mm}
\captionsetup{type=figure}
\caption{Free-form image inpainting results by our system built on gated convolution. Each triad shows original image, free-form input and our result from left to right. The system supports free-form mask and guidance like user sketch. It helps user remove distracting objects, modify image layouts and edit faces in images.}
\label{figs:teaser}
\end{center}
}]

\begin{abstract}
We present a generative image inpainting system to complete images with free-form mask and guidance. The system is based on gated convolutions learned from millions of images without additional labelling efforts. The proposed gated convolution solves the issue of vanilla convolution that treats all input pixels as valid ones, generalizes partial convolution by providing a learnable dynamic feature selection mechanism for each channel at each spatial location across all layers. Moreover, as free-form masks may appear anywhere in images with any shape, global and local GANs designed for a single rectangular mask are not applicable. Thus, we also present a patch-based GAN loss, named SN-PatchGAN, by applying spectral-normalized discriminator on dense image patches. SN-PatchGAN is simple in formulation, fast and stable in training. Results on automatic image inpainting and user-guided extension demonstrate that our system generates higher-quality and more flexible results than previous methods. Our system helps user quickly remove distracting objects, modify image layouts, clear watermarks and edit faces. Code, demo and models are available at: \url{https://github.com/JiahuiYu/generative_inpainting}.
\end{abstract}

\section{Introduction}
Image inpainting (a.k.a.\ image completion or image hole-filling) is a task of synthesizing alternative contents in missing regions such that the modification is visually realistic and semantically correct. It allows to remove distracting objects or retouch undesired regions in photos. It can also be extended to tasks including image/video un-cropping, rotation, stitching, re-targeting, re-composition, compression, super-resolution, harmonization and many others.

In computer vision, two broad approaches to image inpainting exist: patch matching using low-level image features and feed-forward generative models with deep convolutional networks. The former approach~\cite{barnes2009patchmatch, efros2001image, efros1999texture} can synthesize plausible stationary textures, but usually makes critical failures in non-stationary cases like complicated scenes, faces and objects. The latter approach~\cite{iizuka2017globally, yu2018generative, xiong2019foreground, yang2018image, song2018spg, song2018contextual, yu2018wide, nazeri2019edgeconnect, zheng2019pluralistic, sagong2019pepsi, shin2019pepsipp, kim2019deep} can exploit semantics learned from large scale datasets to synthesize contents in non-stationary images in an end-to-end fashion.

However, deep generative models based on vanilla convolutions are naturally ill-fitted for image hole-filling because the spatially shared convolutional filters treat all input pixels or features as same valid ones. For hole-filling, the input to each layer are composed of valid pixels/features outside holes and invalid ones in masked regions. Vanilla convolutions apply same filters on all valid, invalid and mixed (for example, the ones on hole boundary) pixels/features, leading to visual artifacts such as color discrepancy, blurriness and obvious edge responses surrounding holes when tested on free-form masks~\cite{iizuka2017globally, yu2018generative}.

To address this limitation, recently partial convolution~\cite{liu2018image} is proposed where the convolution is masked and normalized to be conditioned only on valid pixels. It is then followed by a rule-based mask-update step to update valid locations for next layer. Partial convolution categorizes all input locations to be either invalid or valid, and multiplies a zero-or-one mask to inputs throughout all layers. The mask can also be viewed as a single un-learnable feature gating channel\footnote{Applying mask before convolution or after is equivalent when convolutions are stacked layer-by-layer in neural networks. Because the output of current layer is the input to next layer and the masked region of input image is already filled with zeros.}. However this assumption has several limitations. First, considering the input spatial locations across different layers of a network, they may include (1) valid pixels in input image, (2) masked pixels in input image, (3) neurons with receptive field covering no valid pixel of input image, (4) neurons with receptive field covering different number of valid pixels of input image (these valid image pixels may also have different relative locations), and (5) synthesized pixels in deep layers. Heuristically categorizing all locations to be either invalid or valid ignores these important information. Second, if we extend to user-guided image inpainting where users provide sparse sketch inside the mask, should these pixel locations be considered as valid or invalid? How to properly update the mask for next layer? Third, for partial convolution the ``invalid'' pixels will progressively disappear layer by layer and the rule-based mask will be all ones in deep layers. However, to synthesize pixels in hole these deep layers may also need the information of whether current locations are inside or outside the hole. The partial convolution with all-ones mask cannot provide such information. We will show that if we allow the network to learn the mask automatically, the mask may have different values based on whether current locations are masked or not in input image, even in deep layers.

We propose gated convolution for free-form image inpainting. It learns a dynamic feature gating mechanism for each channel and each spatial location (for example, inside or outside masks, RGB channels or user-guidance channels). Specifically we consider the formulation where the input feature is firstly used to compute gating values \(g = \sigma(w_gx)\) (\(\sigma\) is sigmoid function, \(w_g\) is learnable parameter). The final output is a multiplication of learned feature and gating values \(y = \phi(wx) \odot g\) where \(\phi\) can be any activation function. Gated convolution is easy to implement and performs significantly better when (1) the masks have arbitrary shapes and (2) the inputs are no longer simply RGB channels with a mask but also have conditional inputs like sparse sketch. For network architectures, we stack gated convolution to form an encoder-decoder network following~\cite{yu2018generative}. Our inpainting network also integrates contextual attention module within same refinement network~\cite{yu2018generative} to better capture long-range dependencies.

Without compromise of performance, we also significantly simplify training objectives as two terms: a pixel-wise reconstruction loss and an adversarial loss. The modification is mainly designed for free-form image inpainting. As the holes may appear anywhere in images with any shape, global and local GANs~\cite{iizuka2017globally} designed for a single rectangular mask are not applicable. Instead, we propose a variant of generative adversarial networks, named SN-PatchGAN, motivated by global and local GANs~\cite{iizuka2017globally}, MarkovianGANs~\cite{li2016precomputed}, perceptual loss~\cite{johnson2016perceptual} and recent work on spectral-normalized GANs~\cite{miyato2018spectral}. The discriminator of SN-PatchGAN directly computes hinge loss on each point of the output map with format \(\mathbb{R}^{h \times w \times c}\), formulating \(h \times w \times c\) number of GANs focusing on different locations and different semantics (represented in different channels). SN-PatchGAN is simple in formulation, fast and stable in training and produces high-quality inpainting results.

\begin{table}[ht]
\centering
\caption{Comparison of different approaches including PatchMatch~\cite{barnes2009patchmatch}, Global\&Local~\cite{iizuka2017globally}, ContextAttention~\cite{yu2018generative}, PartialConv~\cite{liu2018image} and our approach. The comparison of image inpainting is based on four dimensions: Semantic Understanding, Non-Local Algorithm, Free-Form Masks and User-Guided Option.}
\small
\begin{tabular}{@{}l c c c c c@{}} \toprule
 & PM~\cite{barnes2009patchmatch} & GL~\cite{iizuka2017globally} & CA~\cite{yu2018generative} & PC~\cite{liu2018image} & Ours \\
\midrule
Semantics & & \checkmark & \checkmark & \checkmark & \checkmark\\
Non-Local & \checkmark & & \checkmark & &\checkmark\\
Free-Form & \checkmark & & & \checkmark &\checkmark\\
User-guided & \checkmark & & & &\checkmark\\
\bottomrule
\end{tabular}
\label{tabs:intro_summary}
\end{table}

For practical image inpainting tools, enabling user interactivity is crucial because there could exist many plausible solutions for filling a hole in an image. To this end, we present an extension to allow user sketch as guided input. Comparison to other methods is summarized in Table~\ref{tabs:intro_summary}. Our main contributions are as follows: (1) We introduce gated convolution to learn a dynamic feature selection mechanism for each channel at each spatial location across all layers, significantly improving the color consistency and inpainting quality of free-form masks and inputs. (2) We present a more practical patch-based GAN discriminator, SN-PatchGAN, for free-form image inpainting. It is simple, fast and produces high-quality inpainting results. (3) We extend our inpainting model to an interactive one, enabling user sketch as guidance to obtain more user-desired inpainting results. (4) Our proposed inpainting system achieves higher-quality free-form inpainting than previous state of the arts on benchmark datasets including Places2 natural scenes and CelebA-HQ faces. We show that the proposed system helps user quickly remove distracting objects, modify image layouts, clear watermarks and edit faces in images.

\section{Related Work}

\subsection{Automatic Image Inpainting}
A variety of approaches have been proposed for image inpainting. Traditionally, 
patch-based~\cite{efros2001image, efros1999texture} algorithms progressively extend pixels close to the hole boundaries based on low-level features (for example, features of mean square difference on RGB space), to search and paste the most similar image patch. These algorithms work well on stationary textural regions but often fail on non-stationary images. Further, Simakov \etal~propose bidirectional similarity synthesis approach~\cite{simakov2008summarizing} to better capture and summarize non-stationary visual data. To reduce the high cost of memory and computation during search, tree-based acceleration structures of memory~\cite{mount1998ann} and randomized algorithms~\cite{barnes2009patchmatch} are proposed. 
Moreover, inpainting results are improved by matching local features like image gradients~\cite{ballester2001filling, darabi2012image} and offset statistics of similar patches~\cite{he2014image}.
Recently, image inpainting systems based on deep learning are proposed to directly predict pixel values inside masks. A significant advantage of these models is the ability to learn adaptive image features for different semantics. Thus they can synthesize more visually plausible contents especially for images like faces~\cite{li2017generative, yeh2017semantic}, objects~\cite{pathak2016context} and natural scenes~\cite{iizuka2017globally, yu2018generative}. Among all these methods, Iizuka \etal~\cite{iizuka2017globally} propose a fully convolutional image inpainting network with both global and local consistency to handle high-resolution images on a variety of datasets~\cite{karras2017progressive, russakovsky2015imagenet, zhou2017places}. This approach, however, still heavily relies on Poisson image blending with traditional patch-based inpainting results~\cite{he2014image}. Yu \etal~\cite{yu2018generative} propose an end-to-end image inpainting model by adopting stacked generative networks to further ensure the color and texture consistence of generated regions with surroundings. Moreover, for capturing long-range spatial dependencies, contextual attention module~\cite{yu2018generative} is proposed and integrated into networks to explicitly borrow information from distant spatial locations. However, this approach is mainly trained on large rectangular masks and does not generalize well on free-form masks. To better handle irregular masks, partial convolution~\cite{liu2018image} is proposed where the convolution is masked and re-normalized to utilize valid pixels only. It is then followed by a rule-based mask-update step to re-compute new masks layer by layer.

\subsection{Guided Image Inpainting and Synthesis}
To improve image inpainting, user guidance is explored including dots or lines~\cite{ashikhmin2001synthesizing, barnes2009patchmatch, drori2003fragment, sun2005image}, structures~\cite{huang2014image}, transformation or distortion information~\cite{huang2013transformation, pavic2006interactive} and image exemplars~\cite{criminisi2004region, hays2007scene, kwatra2005texture, whyte2009get, zhao2019guided}. Notably, Hays and Efros~\cite{hays2007scene} first utilize millions of photographs as a database to search for an example image which is most similar to the input, and then complete the image by cutting and pasting the corresponding regions from the matched image.

Recent advances in conditional generative networks empower user-guided image processing, synthesis and manipulation learned from large-scale datasets. Here we selectively review several related work. Zhang \etal~\cite{zhang2017real} propose colorization networks which can take user guidance as additional inputs. Wang \etal~\cite{wang2018high} propose to synthesize high-resolution photo-realistic images from semantic label maps using conditional generative adversarial networks. The Scribbler~\cite{sangkloy2017scribbler} explore a deep generative network conditioned on sketched boundaries and sparse color strokes to synthesize cars, bedrooms, or faces.

\subsection{Feature-wise Gating}
Feature-wise gating has been explored widely in vision~\cite{hu2018squeeze, oord2016conditional, srivastava2015highway, wang2017gated}, language~\cite{dauphin2017language}, speech~\cite{oord2016wavenet} and many other tasks. For examples, Highway Networks~\cite{srivastava2015highway} utilize feature gating to ease gradient-based training of very deep networks. Squeeze-and-Excitation Networks re-calibrate feature responses by explicitly multiplying each channel with learned sigmoidal gating values. WaveNets~\cite{oord2016wavenet} achieve better results by employing a special feature gating \(y = \text{tanh}(w_1 ∗ x) \cdot \text{sigmoid}(w_2 ∗ x)\) for modeling audio signals.

\section{Approach}
\begin{figure*}[ht]
  \centering
  \includegraphics[width=\linewidth]{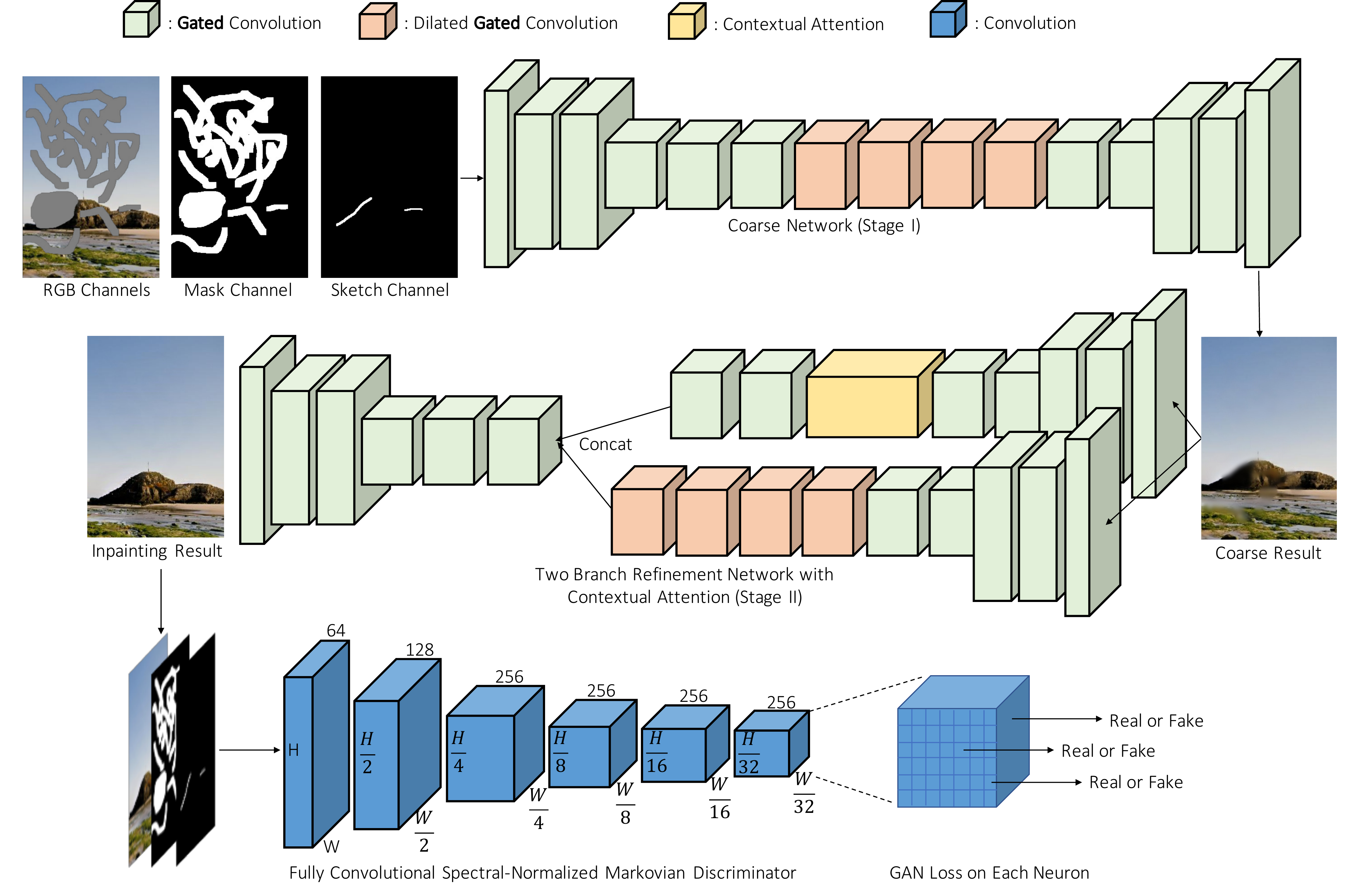}
  \vspace*{-6mm}
  \caption{Overview of our framework with gated convolution and SN-PatchGAN for free-form image inpainting.}
  \label{figs:pipeline}
\end{figure*}

In this section, we describe our approach from bottom to top. We first introduce the details of the Gated Convolution, SN-PatchGAN, and then present the overview of inpainting network in Figure~\ref{figs:pipeline} and our extension to allow optional user guidance.

\subsection{Gated Convolution}
We first explain why vanilla convolutions used in~\cite{iizuka2017globally,yu2018generative} are ill-fitted for the task of free-form image inpainting. We consider a convolutional layer in which a bank of filters are applied to the input feature map as output. Assume input is \(C\mathit{-channel}\), each pixel located at \((y, x)\) in \(C'\mathit{-channel}\) output map is computed as 
\[
    O_{y, x} = \sum_{i = -k_h'}^{k_h'}\sum_{j = -k_w'}^{k_w'} W_{k_h' + i, k_w' + j} \cdot I_{y+i, x+j},
\]
where \(x, y\) represents x-axis, y-axis of output map, \(k_h\) and \(k_w\) is the kernel size (\eg~\(3 \times 3\)), \(k_h' = \frac{k_h - 1}{2}\), \(k_w' = \frac{k_w-1}{2}\), \(W \in \mathbb{R}^{k_h \times k_w \times C' \times C}\) represents convolutional filters, \(I_{y+i, x+j} \in \mathbb{R}^{C}\) and \(O_{y, x} \in \mathbb{R}^{C'}\) are inputs and outputs. For simplicity, the bias in convolution is ignored.

The equation shows that for all spatial locations \((y, x)\), the same filters are applied to produce the output in vanilla convolutional layers. This makes sense for tasks such as image classification and object detection, where all pixels of input image are valid, to extract local features in a sliding-window fashion. However, for image inpainting, the input are composed of both regions with valid pixels/features outside holes and invalid pixels/features (in shallow layers) or synthesized pixels/features (in deep layers) in masked regions. This causes ambiguity during training and leads to visual artifacts such as color discrepancy, blurriness and obvious edge responses during testing, as reported in~\cite{liu2018image}.

Recently partial convolution is proposed~\cite{liu2018image} which adapts a masking and re-normalization step to make the convolution dependent only on valid pixels as
\[
    O_{y, x} = \left\{
    \begin{array}{@{}ll@{}}
        \sum \sum W \cdot (I \odot \frac{M}{\mathit{sum}(M)}), & \text{if sum(M) $>$ 0}\\
        0, & \text{otherwise}
    \end{array}\right.
\]
in which \(M\) is the corresponding binary mask, \(1\) represents pixel in the location \((y, x)\) is valid, \(0\) represents the pixel is invalid, \(\odot\) denotes element-wise multiplication. After each partial convolution operation, the mask-update step is required to propagate new \(M\) with the following rule: \(m'_{y, x} = 1, \text{iff sum(M) $>$ 0}\).

Partial convolution~\cite{liu2018image} improves the quality of inpainting on irregular mask, but it still has remaining issues: (1) It heuristically classifies all spatial locations to be either valid or invalid. The mask in next layer will be set to ones no matter how many pixels are covered by the filter range in previous layer (for example, 1 valid pixel and 9 valid pixels are treated as same to update current mask). (2) It is incompatible with additional user inputs. We aim at a user-guided image inpainting system where users can optionally provide sparse sketch inside the mask as conditional channels. In this situation, should these pixel locations be considered as valid or invalid? How to properly update the mask for next layer? (3) For partial convolution the invalid pixels will progressively disappear in deep layers, gradually converting all mask values to ones. However, our study shows that if we allow the network to learn optimal mask automatically, the network assigns soft mask values to every spatial locations even in deep layers. (4) All channels in each layer share the same mask, which limits the flexibility. Essentially, partial convolution can be viewed as un-learnable single-channel feature hard-gating.

\begin{figure}[h]
  \centering
  \includegraphics[width=0.9\linewidth]{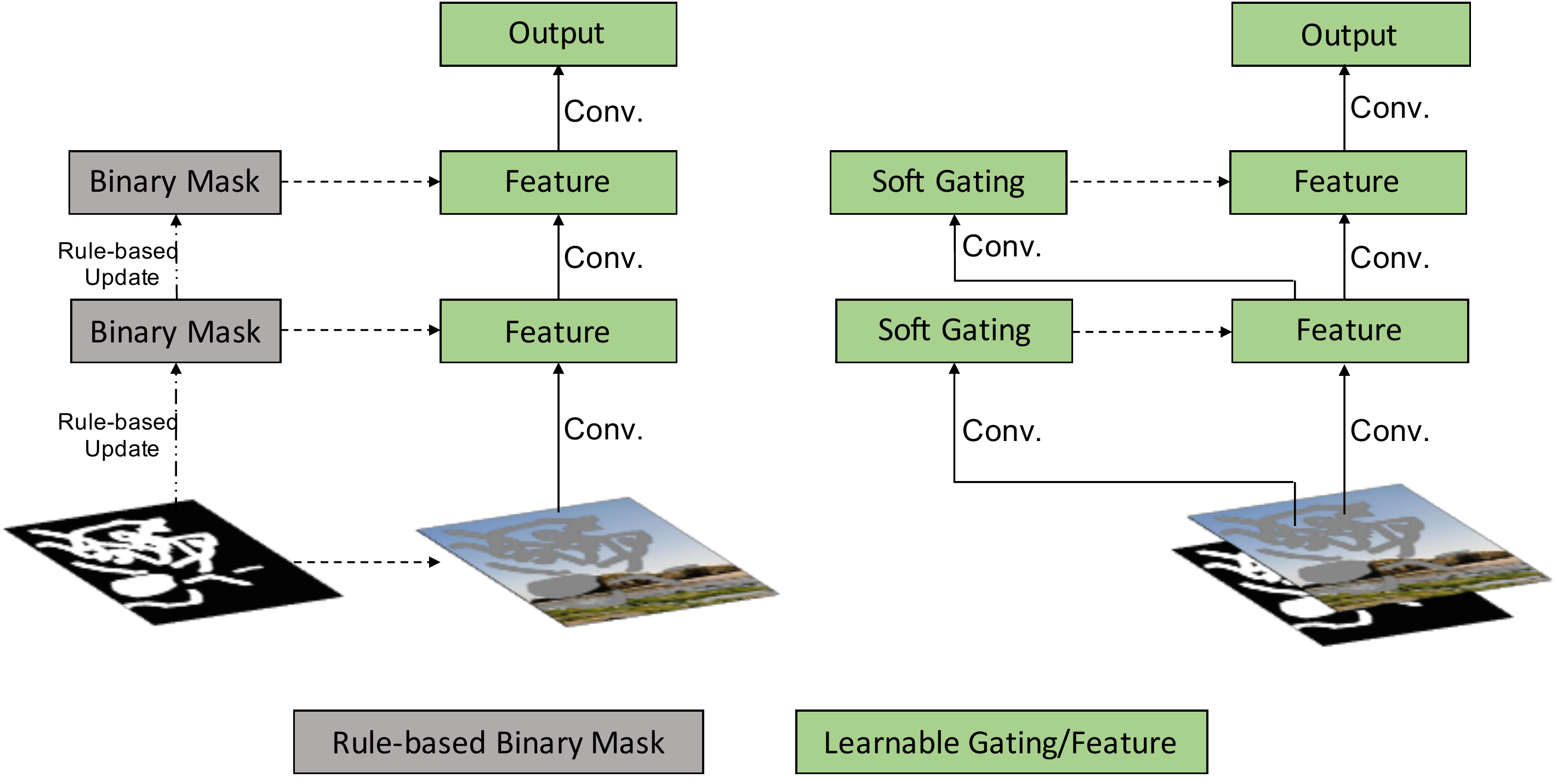}
  \caption{Illustration of partial convolution (left) and gated convolution (right).}
  \label{figs:gated_conv}
\end{figure}

We propose gated convolution for image inpainting network, as shown in Figure~\ref{figs:gated_conv}. Instead of hard-gating mask updated with rules, gated convolutions learn soft mask automatically from data. It is formulated as:
\[
    \begin{split}
        \textit{Gating}_{y, x} & =  \sum \sum W_g \cdot I\\
        \textit{Feature}_{y, x} & =  \sum \sum W_f \cdot I\\
        O_{y, x} & = \phi(\textit{Feature}_{y, x}) \odot \sigma(\textit{Gating}_{y, x})
    \end{split}
\]
where \(\sigma\) is sigmoid function thus the output gating values are between zeros and ones. \(\phi\) can be any activation functions (for examples, ReLU, ELU and LeakyReLU). \(W_g\) and \(W_f\) are two different convolutional filters.

The proposed gated convolution learns a dynamic feature selection mechanism for each channel and each spatial location. Interestingly, visualization of intermediate gating values show that it learns to select the feature not only according to background, mask, sketch, but also considering semantic segmentation in some channels. Even in deep layers, gated convolution learns to highlight the masked regions and sketch information in separate channels to better generate inpainting results.

\subsection{Spectral-Normalized Markovian Discriminator (SN-PatchGAN)}
For previous inpainting networks which try to fill a single rectangular hole, an additional local GAN is used on the masked rectangular region to improve results~\cite{iizuka2017globally, yu2018generative}. However, we consider the task of free-form image inpainting where there may be multiple holes with any shape at any location. Motivated by global and local GANs~\cite{iizuka2017globally}, MarkovianGANs~\cite{isola2017image, li2016precomputed}, perceptual loss~\cite{johnson2016perceptual} and recent work on spectral-normalized GANs~\cite{miyato2018spectral}, we present a simple and effective GAN loss, SN-PatchGAN, for training free-form image inpainting networks. 

A convolutional network is used as the discriminator where the input consists of image, mask and guidance channels, and the output is a 3-D feature of shape \(\mathbb{R}^{h \times w \times c}\) (\(h\), \(w\), \(c\) representing the height, width and number of channels respectively). As shown in Figure~\ref{figs:pipeline}, six strided convolutions with kernel size \(5\) and stride \(2\) is stacked to captures the feature statistics of Markovian patches~\cite{li2016precomputed}. We then directly apply GANs for each feature element in this feature map, formulating \(h \times w \times c\) number of GANs focusing on different locations and different semantics (represented in different channels) of input image. It is noteworthy that the receptive field of each neuron in output map can cover entire input image in our training setting, thus a global discriminator is not necessary.

We also adapt the recently proposed spectral normalization~\cite{miyato2018spectral} to further stabilize the training of GANs. We use the default fast approximation algorithm of spectral normalization described in SN-GANs~\cite{miyato2018spectral}. To discriminate if the input is real or fake, we also use the hinge loss as objective function for generator \(
\mathcal{L}_{G} = - \mathbb{E}_{z \sim \mathbb{P}_{z}(z)} [D^{sn}(G(z))]
\) and discriminator \(
    \mathcal{L}_{D^{sn}} = \mathbb{E}_{x \sim \mathbb{P}_{data}(x)} [ ReLU(\mathbbm{1}-D^{sn}(x))] + \mathbb{E}_{z \sim \mathbb{P}_{z}(z)}[ ReLU(\mathbbm{1}+D^{sn}(G(z)))]
\) where \(D^{sn}\) represents spectral-normalized discriminator, \(G\) is image inpainting network that takes incomplete image \(z\).

With SN-PatchGAN, our inpainting network trains faster and more stable than baseline model~\cite{yu2018generative}. Perceptual loss is not used since similar patch-level information is already encoded in SN-PatchGAN. Compared with PartialConv~\cite{liu2018image} in which \(6\) different loss terms and balancing hyper-parameters are used, our final objective function for inpainting network is only composed of pixel-wise \(\ell_1\) reconstruction loss and SN-PatchGAN loss, with default loss balancing hyper-parameter as \(1:1\).

\subsection{Inpainting Network Architecture}
We customize a generative inpainting network~\cite{yu2018generative} with the proposed gated convolution and SN-PatchGAN loss. Specifically, we adapt the full model architecture in~\cite{yu2018generative} with both coarse and refinement networks. The full framework is summarized in Figure~\ref{figs:pipeline}.

For coarse and refinement networks, we use a simple encoder-decoder network~\cite{yu2018generative} instead of U-Net used in PartialConv~\cite{liu2018image}. We found that skip connections in a U-Net~\cite{ronneberger2015u} have no significant effect for non-narrow mask. This is mainly because for center of a masked region, the inputs of these skip connections are almost zeros thus cannot propagate detailed color or texture information to the decoder of that region. For hole boundaries, our encoder-decoder architecture equipped with gated convolution is sufficient to generate seamless results.

We replace all vanilla convolutions with gated convolutions~\cite{yu2018generative}. One potential problem is that gated convolutions introduce additional parameters. To maintain the same efficiency with our baseline model~\cite{yu2018generative}, we slim the model width by \(25\%\) and have not found obvious performance drop both quantitatively and qualitatively. The inpainting network is trained end-to-end and can be tested on free-form holes at arbitrary locations. Our network is fully convolutional and supports different input resolutions in inference.

\subsection{Free-Form Mask Generation}
The algorithm to automatically generate free-form masks is important and non-trivial. The sampled masks, in essence, should be (1) similar to masks drawn in real use-cases, (2) diverse to avoid over-fitting, (3) efficient in computation and storage, (4) controllable and flexible. Previous method~\cite{liu2018image} collects a fixed set of irregular masks from an occlusion estimation method between two consecutive frames of videos. Although random dilation, rotation and cropping are added to increase its diversity, the method does not meet other requirements listed above.

We introduce a simple algorithm to automatically generate random free-form masks on-the-fly during training. For the task of hole filling, users behave like using an eraser to brush back and forth to mask out undesired regions. This behavior can be simply simulated with a randomized algorithm by drawing lines and rotating angles repeatedly. To ensure smoothness of two lines, we also draw a circle in joints between the two lines. More details are included in the supplementary materials  due to space limit.

\subsection{Extension to User-Guided Image Inpainting}
We use sketch as an example user guidance to extend our image inpainting network as a user-guided system. Sketch (or edge) is simple and intuitive for users to draw. We show both cases with faces and natural scenes. For faces, we extract landmarks and connect related landmarks. For natural scene images, we directly extract edge maps using the HED edge detector~\cite{xie2015holistically} and set all values above a certain threshold (\ie~0.6) to ones. Sketch examples are shown in the supplementary materials due to space limit.

For training the user-guided image inpainting system, intuitively we will need additional constraint loss to enforce the network generating results conditioned on the user guidance. However with the same combination of pixel-wise reconstruction loss and GAN loss (with conditional channels as input to the discriminator), we are able to learn conditional generative network in which the generated results respect user guidance faithfully. We also tried to use additional pixel-wise loss on HED~\cite{xie2015holistically} output features with the raw image or the generated result as input to enforce constraints, but the inpainting quality is similar. The user-guided inpainting model is separately trained with a 5-channel input (R,G,B color channels, mask channel and sketch channel).

\section{Results}
\label{secs:results}

We evaluate the proposed free-form image inpainting system on Places2~\cite{zhou2017places} and CelebA-HQ faces~\cite{karras2017progressive}.
Our model has totally 4.1M parameters, and is trained with TensorFlow v1.8, CUDNN v7.0, CUDA v9.0.
For testing, it runs at 0.21 seconds per image on single NVIDIA(R) Tesla(R) V100 GPU and 1.9 seconds on Intel(R) Xeon(R) CPU @ 2.00GHz for images of resolution \(512 \times 512\) on average, regardless of hole size.

\subsection{Quantitative Results}
As mentioned in~\cite{yu2018generative}, image inpainting lacks good quantitative evaluation metrics.
Nevertheless, we report in Table~\ref{tabs:quantitative} our evaluation results in terms of mean \(\ell_1\) error and mean \(\ell_2\) error on validation images of Places2, with both center rectangle mask and free-form mask.
As shown in the table, learning-based methods perform better than PatchMatch~\cite{barnes2009patchmatch} in terms of mean \(\ell_1\) and \(\ell_2\) errors.
Moreover, partial convolution implemented within the same framework obtains worse performance, which may due to un-learnable rule-based gating. %
\begin{table}[ht]
\centering
\caption{Results of mean \(\ell_1\) error and mean \(\ell_2\) error on validation images of Places2 with both rectangle masks and free-form masks. Both PartialConv* and ours are trained on same random combination of rectangle and free-form masks. No edge guidance is utilized in training/inference to ensure fair comparison. * denotes our implementation within the same framework due to unavailability of official implementation and models.}
\small
\begin{tabular}{@{}l c c l c c@{}} \toprule
 & \multicolumn{2}{c}{rectangular mask} && \multicolumn{2}{c}{free-form mask}\\
\cmidrule{2-3} \cmidrule{5-6}
Method & \(\ell_1\) err. & \(\ell_2\) err. && \(\ell_1\) err. & \(\ell_2\) err. \\
\midrule
PatchMatch~\cite{barnes2009patchmatch} & 16.1\% & 3.9\% && 11.3\% & 2.4\% \\
Global\&Local~\cite{iizuka2017globally} & 9.3\% & 2.2\% && 21.6\% & 7.1\%\\
ContextAttention~\cite{yu2018generative} & \textbf{8.6\%} & 2.1\% && 17.2\% & 4.7\%\\
PartialConv*~\cite{liu2018image} & 9.8\% & 2.3\% && 10.4\% & 1.9\%\\
Ours & \textbf{8.6\%} & \textbf{2.0\%} && \textbf{9.1\%} & \textbf{1.6\%}\\
\bottomrule
\end{tabular}
\label{tabs:quantitative}
\end{table}

\subsection{Qualitative Comparisons}
\begin{figure*}[t]
  \centering
  \includegraphics[width=\linewidth]{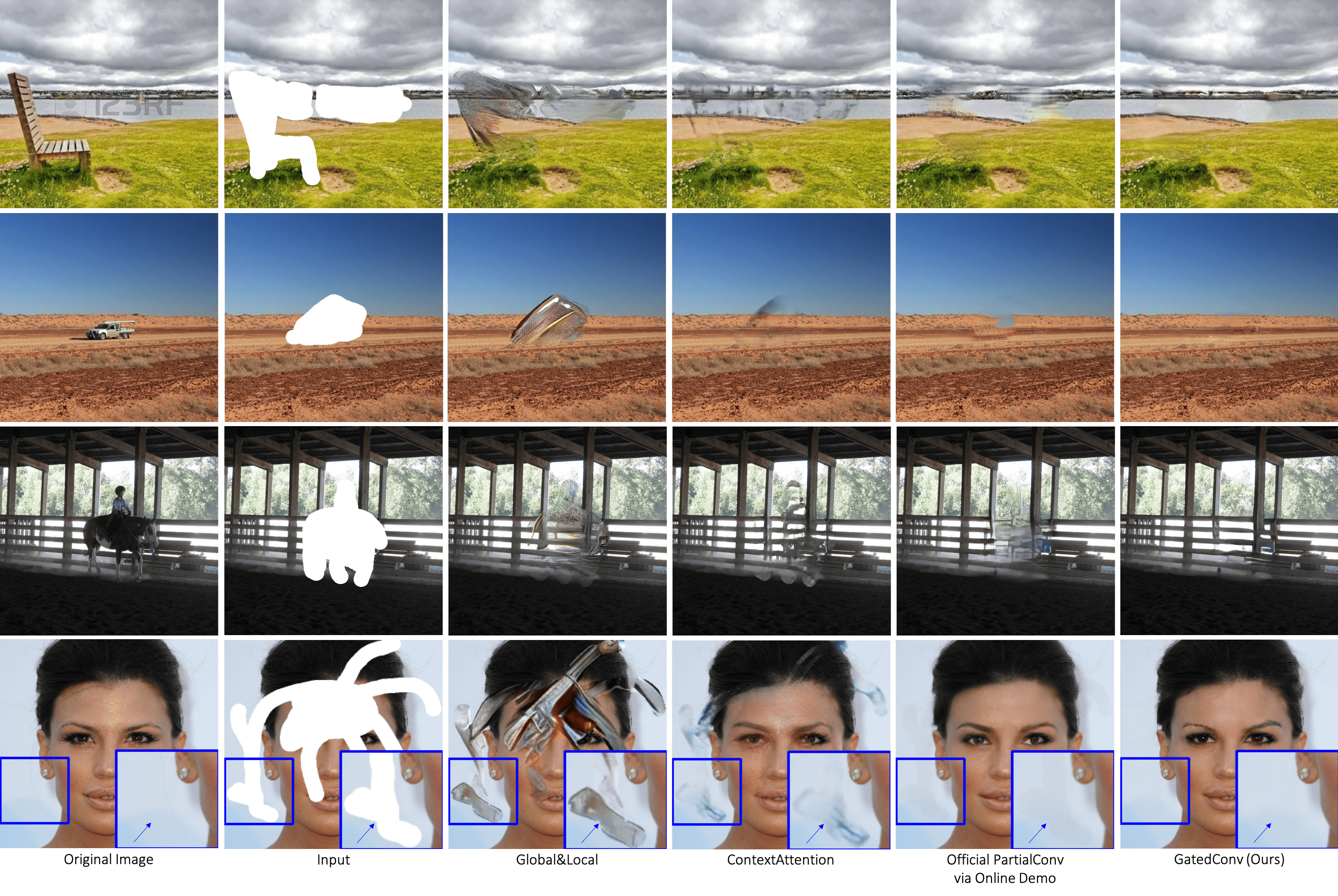}
  \vspace*{-5mm}
  \caption{Example cases of qualitative comparison on the Places2 and CelebA-HQ validation sets. More comparisons are included in supplementary materials due to space limit. Best viewed (\eg, shadows in uniform region) with zoom-in.}
  \label{figs:comparisons}
\end{figure*}

\begin{figure}[t]
  \centering
  \includegraphics[width=\linewidth]{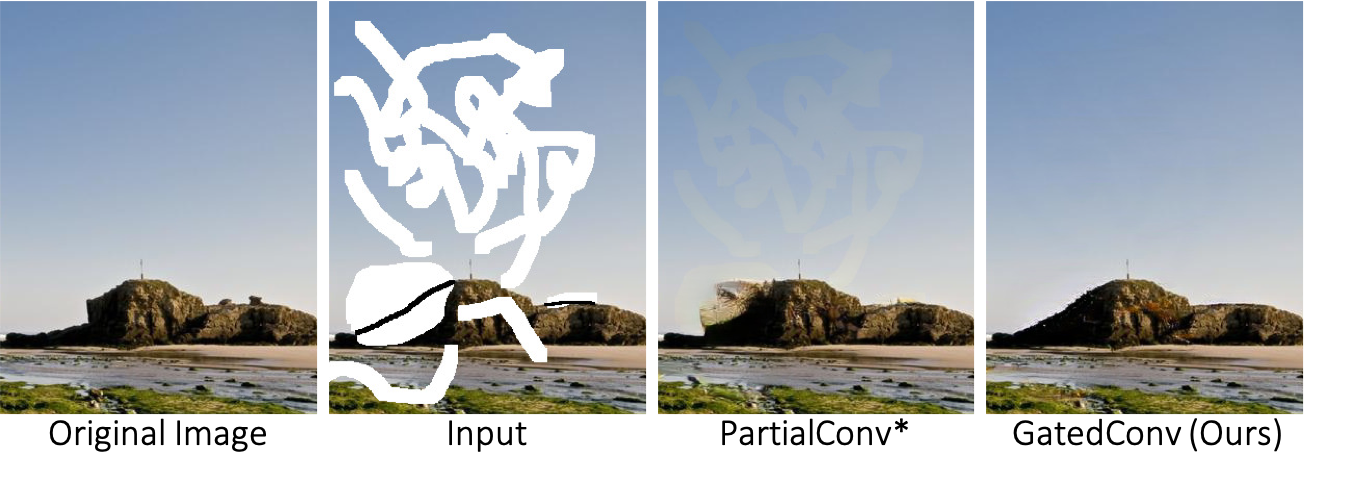}
  \vspace{-8mm}
  \caption{Comparison of user-guided image inpainting.}
  \label{figs:guided_comparisons}
  \vspace{-2mm}
\end{figure}

Next, we compare our model with previous state-of-the-art methods~\cite{iizuka2017globally, liu2018image, yu2018generative}.
Figure~\ref{figs:comparisons} and Figure~\ref{figs:guided_comparisons} shows automatic and user-guided inpainting results with several representative images.
For automatic image inpainting, the result of PartialConv is obtained from its online demo\footnote{https://www.nvidia.com/research/inpainting}.
For user-guided image inpainting, we train PartialConv* with the exact same setting of GatedConv, expect the convolution types (sketch regions are treated as valid pixels for rule-based mask updating). For all learning-based methods, no post-processing step is performed to ensure fairness.

\begin{figure*}[ht]
  \centering
  \vspace{-6mm}
  \includegraphics[width=\linewidth]{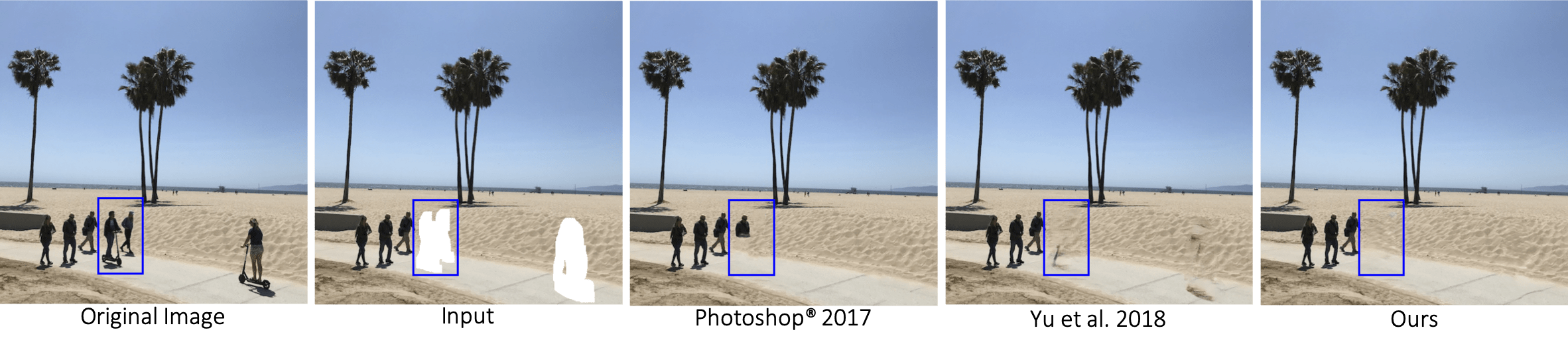}
  \vspace*{-8mm}
  \caption{Object removal case study with comparison.}
  \label{figs:object_removal}
\end{figure*}

As reported in~\cite{iizuka2017globally}, simple uniform region (last row of Figure~\ref{figs:comparisons} and Figure~\ref{figs:guided_comparisons}) are hard cases for learning-based image inpainting networks.
Previous methods with vanilla convolution have obvious visual artifacts and edge responses in/surrounding holes.
PartialConv produces better results but still exhibits observable color discrepancy.
Our method based on gated convolution obtains more visually pleasing results without noticeable color inconsistency.
In Figure~\ref{figs:guided_comparisons}, given sparse sketch, our method produces realistic results with seamless boundary transitions.

\subsection{Object Removal and Creative Editing}
Moreover, we study two important real use cases of image inpainting: object removal and creative editing.

\textbf{Object Removal.}
In the first example, we try to remove the distracting person in Figure~\ref{figs:object_removal}.
We compare our method with commercial product Photoshop (based on PatchMatch~\cite{barnes2009patchmatch}) and the previous state-of-the-art generative inpainting network (official released model trained on Places2)~\cite{yu2018generative}.
The results show that \textit{Content-Aware Fill} function from Photoshop incorrectly copies half of face from left.
This example reflects the fact that traditional methods without learning from large-scale data ignore the semantics of an image, which leads to critical failures in non-stationary/complicated scenes.
For learning-based methods with vanilla convolution~\cite{yu2018generative}, artifacts exist near hole boundaries.

\begin{figure}[t]
  \centering
  \includegraphics[width=\linewidth]{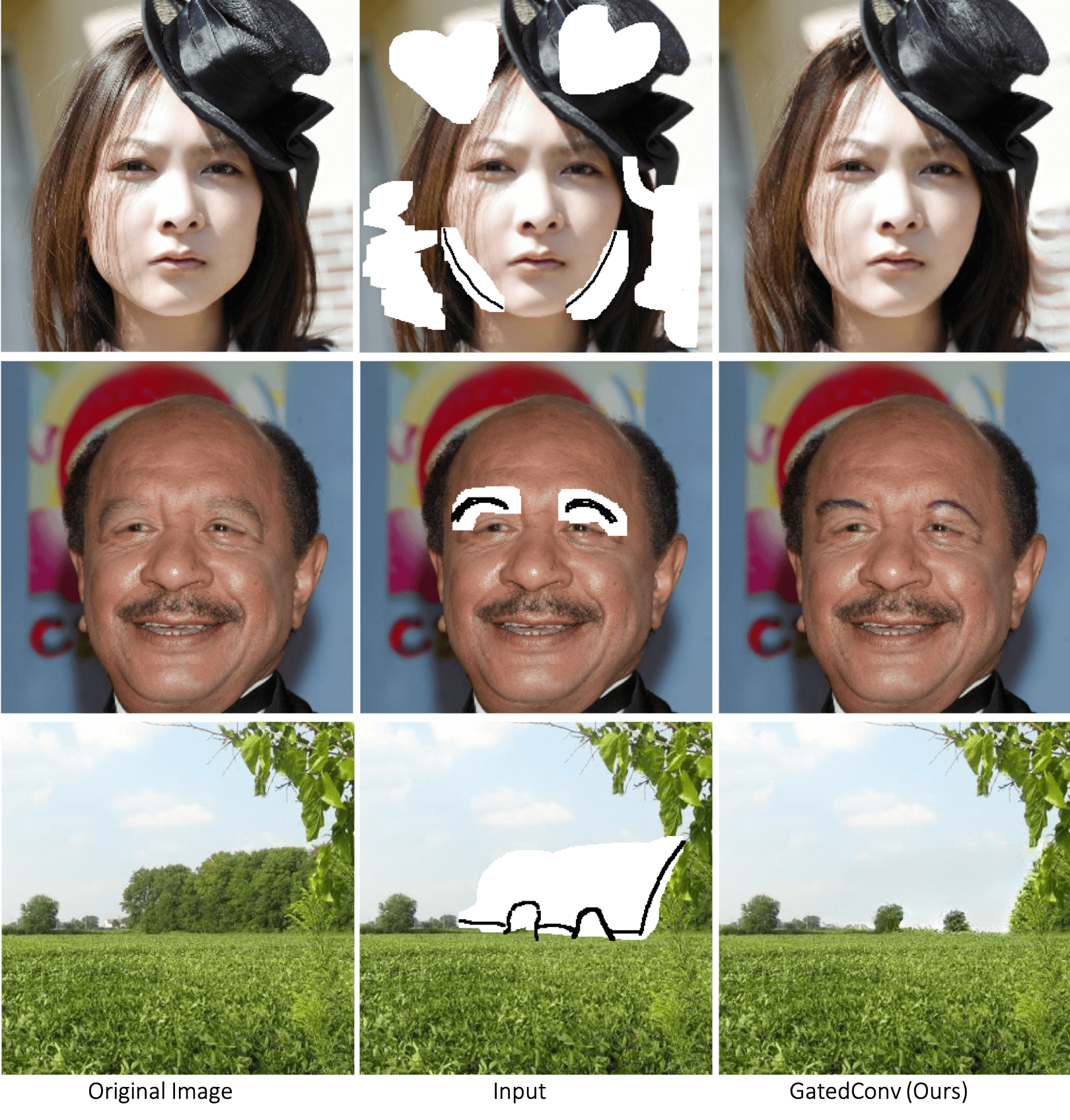}
  \vspace{-8mm}
  \caption{Examples of user-guided inpainting/editing of faces and natural scenes.}
  \label{figs:creative_editing}
  \vspace{-2mm}
\end{figure}

\textbf{Creative Editing.}
Next we study the case where user interacts with the inpainting system to produce more desired results. The examples on both faces and natural scenes are shown in Figure~\ref{figs:creative_editing}. Our inpainting results nicely follow the user sketch, which is useful for creatively editing image layouts, faces and many others.

\subsection{User Study}
We performed a user study by first collecting 30 test images (with holes but no sketches) from Places2 validation dataset without knowing their inpainting results on each model.
We then computed results of the following four methods for comparison: (1) ground truth, (2) our model, (3) re-implemented PartialConv~\cite{liu2018image} within same framework, and (4) official PartialConv~\cite{liu2018image}.
We did two types of user study.
(A) We evaluate each method individually to rate the naturalness/inpainting quality of results (from 1 to 10, the higher the better), and (B) we compare our model and the official PartialConv model to evaluate which method produces better results.
104 users finished the user study with the results shown as follows.

(A) Naturalness: (1) 9.89, (2) 7.72, (3) 7.07, (4) 6.54

(B) Pairwise comparison of (2) our model vs. (4) official PartialConv model: 79.4\% vs. 20.6\% (the higher the better).

\subsection{Ablation Study of SN-PatchGAN}
\begin{figure}[ht]
  \centering
  \includegraphics[width=\linewidth]{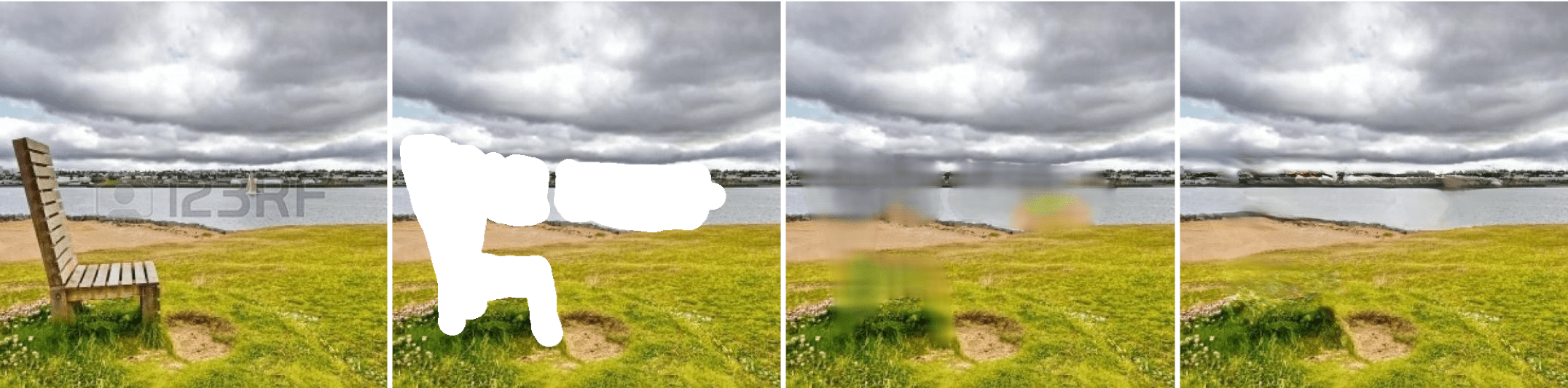}
  \vspace{-6mm}
  \caption{Ablation study of SN-PatchGAN. From left to right, we show original image, masked input, results with one global GAN and our results with SN-PatchGAN.}
  \label{figs:gan_ablation}
\end{figure}

SN-PatchGAN is proposed for the reason that free-form masks may appear anywhere in images with any shape.
Previously introduced global and local GANs~\cite{iizuka2017globally} designed for a single rectangular mask are not applicable.
We provide ablation experiments of SN-PatchGAN in the context of image inpainting in Figure~\ref{figs:gan_ablation}.
SN-PatchGAN leads to significantly better results, which verifies that (1) one vanilla global discriminator has worse performance~\cite{iizuka2017globally}, and (2) GAN with spectral normalization has better stability and performance~\cite{miyato2018spectral}. Although introducing more loss functions may help in training free-form image inpainting networks~\cite{liu2018image}, we demonstrate that a simple combination of SN-PatchGAN loss and pixel-wise \(\ell_1\) loss, with default loss balancing hyper-parameter as \num{1}:\num{1}, produces photo-realistic inpainting results. More comparison examples are shown in the supplementary materials.

\section{Conclusions}
We presented a novel free-form image inpainting system based on an end-to-end generative network with gated convolution, trained with pixel-wise \(\ell_1\) loss and SN-PatchGAN. We demonstrated that gated convolutions significantly improve inpainting results with free-form masks and user guidance input. We showed user sketch as an exemplar guidance to help users quickly remove distracting objects, modify image layouts, clear watermarks, edit faces and interactively create novel objects in photos. Quantitative results, qualitative comparisons and user studies demonstrated the superiority of our proposed free-form image inpainting system.

{\small
\bibliographystyle{ieee_fullname}
\bibliography{egbib}
}

\newpage
\appendix

In this supplementary material, we first provide details of our free-form mask generation algorithm in Section~\ref{secs:mask} and sketch generation algorithm in Section~\ref{secs:sketch}.
We then study the effects of sketch input in Section~\ref{secs:sketch_effect} with an example where the input image uses the same mask but different sketches.
Next we provide visualization and interpretation of learned gating values in Section~\ref{secs:vis_network}.
We show additional ablation study of our proposed SN-PatchGAN in Section~\ref{secs:ablation}.
We show more comparison results of Global\&Local~\cite{iizuka2017globally}, ContextAttention~\cite{yu2018generative}, PartialConv~\cite{liu2018image} (both our implementation within same framework and official model via online demo\footnote{https://www.nvidia.com/research/inpainting/}) and our GatedConv in Section~\ref{secs:comparison}. We finally show more inpainting results of our system with support of free-form masks and user guidance on both natural scenes and faces in Section~\ref{secs:more_results}. Moreover, a recorded \textit{real-time} video demo is available at: \url{https://youtu.be/uZkEi9Y2dj4}.

\section{Free-Form Mask Generation} \label{secs:mask}
\begin{figure}[h]
  \centering
  \includegraphics[width=\linewidth]{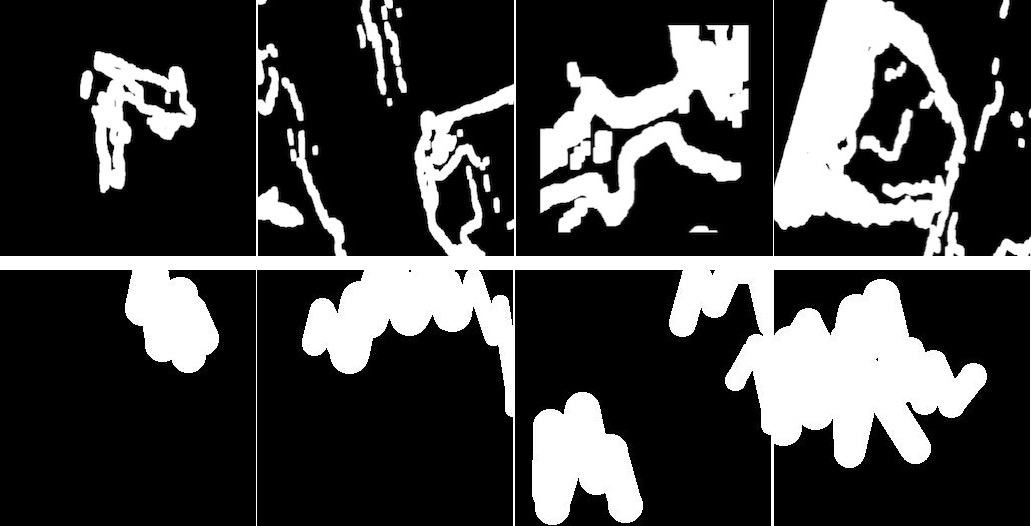}
  \caption{Sampled free-form masks with previous work~\cite{liu2018image} (1st row) and our automatic algorithm (2nd row). }
  \label{figs:masks}
\end{figure}

The algorithm to automatically generate free-form masks is important and non-trivial. The sampled masks, in essence, should be (1) similar in shape to holes drawn in real use-cases, (2) diverse to avoid over-fitting, (3) efficient in computation and storage, (4) controllable and flexible. Previous method~\cite{liu2018image} collects a fixed set of irregular masks from an occlusion estimation method between two consecutive frames of videos. Although random dilation, rotation and cropping are added to increase its diversity, the method does not meet other requirements listed above. 

We introduce a simple algorithm to automatically generate random free-form masks on-the-fly during training. For the task of hole filling, users behave like using an eraser to brush back and forth to mask out undesired regions. This behavior can be simply simulated with a randomized algorithm by drawing lines and rotating angles repeatedly. To ensure smoothness of two lines, we also draw a circle in joints between the two lines.

\begin{algorithm}
  \caption{Algorithm for sampling free-form training masks. \textit{maxVertex}, \textit{maxLength}, \textit{maxBrushWidth}, \textit{maxAngle} are four hyper-parameters to control the mask generation.}
  \label{algos:masks}
  \begin{algorithmic}
    \State mask = zeros(imageHeight, imageWidth)
    \State numVertex = random.uniform(maxVertex)
    \State startX = random.uniform(imageWidth)
    \State startY = random.uniform(imageHeight)
    \State brushWidth = random.uniform(maxBrushWidth)
    \For  {$i=0$ to numVertex}
    \State angle = random.uniform(maxAngle)
    \If{  (i \% 2 == 0) }
    \State angle = 2 * pi - angle // comment: reverse mode
    \EndIf  
    \State length = random.uniform(maxLength)
    \State Draw line from point (startX, startY) with angle, length and brushWidth as line width.
    \State startX = startX + length * sin(angle)
    \State startY = startY + length * cos(angle)
    \State Draw a circle at point (startX, startY) with radius as half of brushWidth. // comment: ensure smoothness of strokes.
    \EndFor
    \State mask = random.flipLeftRight(mask)
    \State mask = random.flipTopBottom(mask)
  \end{algorithmic}
\end{algorithm}

We use \textit{maxVertex}, \textit{maxLength}, \textit{maxWidth} and \textit{maxAngle} as four hyper-parameters to provide large varieties of sampled masks. Moreover, our algorithm generates masks on-the-fly with little computational overhead and no storage is required. In practice, the computation of free-form masks on CPU can be easily hid behind training networks on GPU in modern deep learning frameworks. The overall mask generation algorithm is illustrated in Algorithm~\ref{algos:masks}. Additionally we can sample multiple strokes in single image to mask multiple regions, and add regular masks (\eg~rectangular) on top of sampled free-form masks. Example masks compared with previous method~\cite{liu2018image} is shown in Figure~\ref{figs:masks}.

\section{Sketch Generation} \label{secs:sketch}
\begin{figure}[ht]
  \centering
  \includegraphics[width=\linewidth]{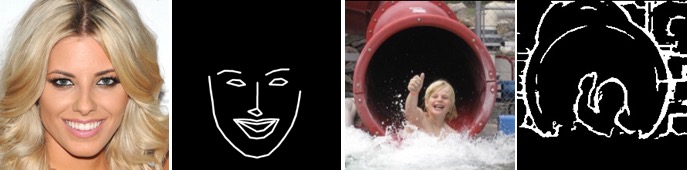}
  \caption{For face dataset (on the left), we directly detect landmarks of faces and connect related nearby landmarks as training sketch, which is extremely robust and useful for editing faces. We use HED~\cite{xie2015holistically} model with threshold \(0.6\) to extract binary sketch for natural scenes (on the right).}
  \label{figs:sketch}
\end{figure}
We use sketch as an example user guidance to extend our image inpainting network as a user guided system. We show both cases on faces and natural scenes. For faces, we extract landmarks and connect related landmarks. For natural scene images, we directly extract edge maps using the HED~\cite{xie2015holistically} edge detector and set all values above a certain threshold (\ie~0.6) to ones. Sketch examples are shown in Figure~\ref{figs:sketch}. Alternative methods to generative better sketch or other user guidance should also work well with our user-guided image inpainting system.

\section{The Effects of Sketch Input}
\label{secs:sketch_effect}
\begin{figure}
\centering
\vspace{-3mm}
\includegraphics[width=\linewidth]{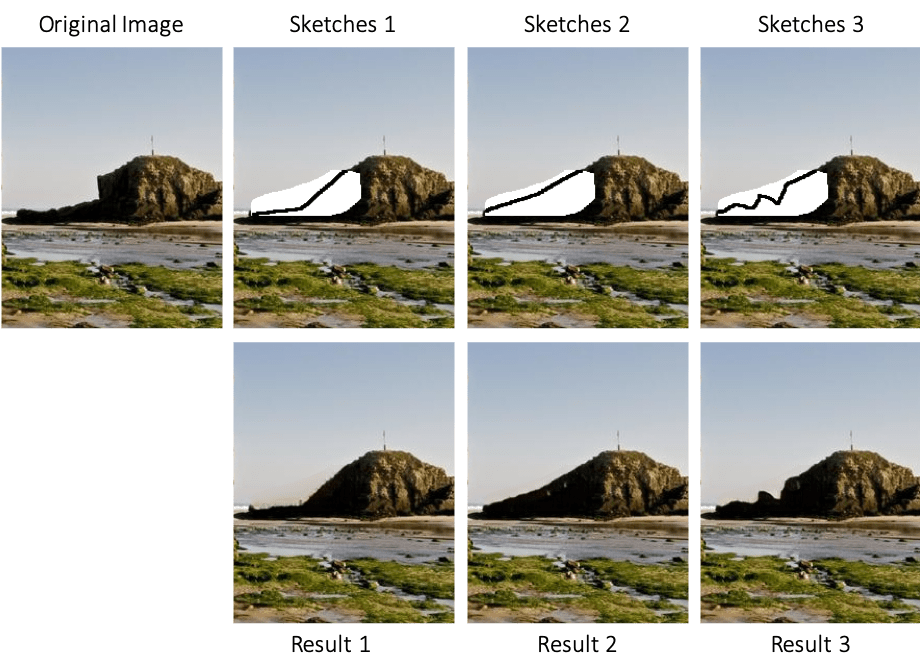}
\caption{Image inpainting examples where the input image uses same mask but different sketches.}
\label{figs:rebuttal}
\vspace{-3mm}
\end{figure}

As shown in Section 4.3, our inpainting network can nicely follow the user sketch, which is useful for creative editing of images. We show in Figure~\ref{figs:rebuttal} an additional comparison case where the input image uses the same mask but different sketches.

\section{Visualization and Interpretation} \label{secs:vis_network}
\begin{figure*}[ht]
  \centering
  \vspace*{-1cm}
  \includegraphics[width=\linewidth]{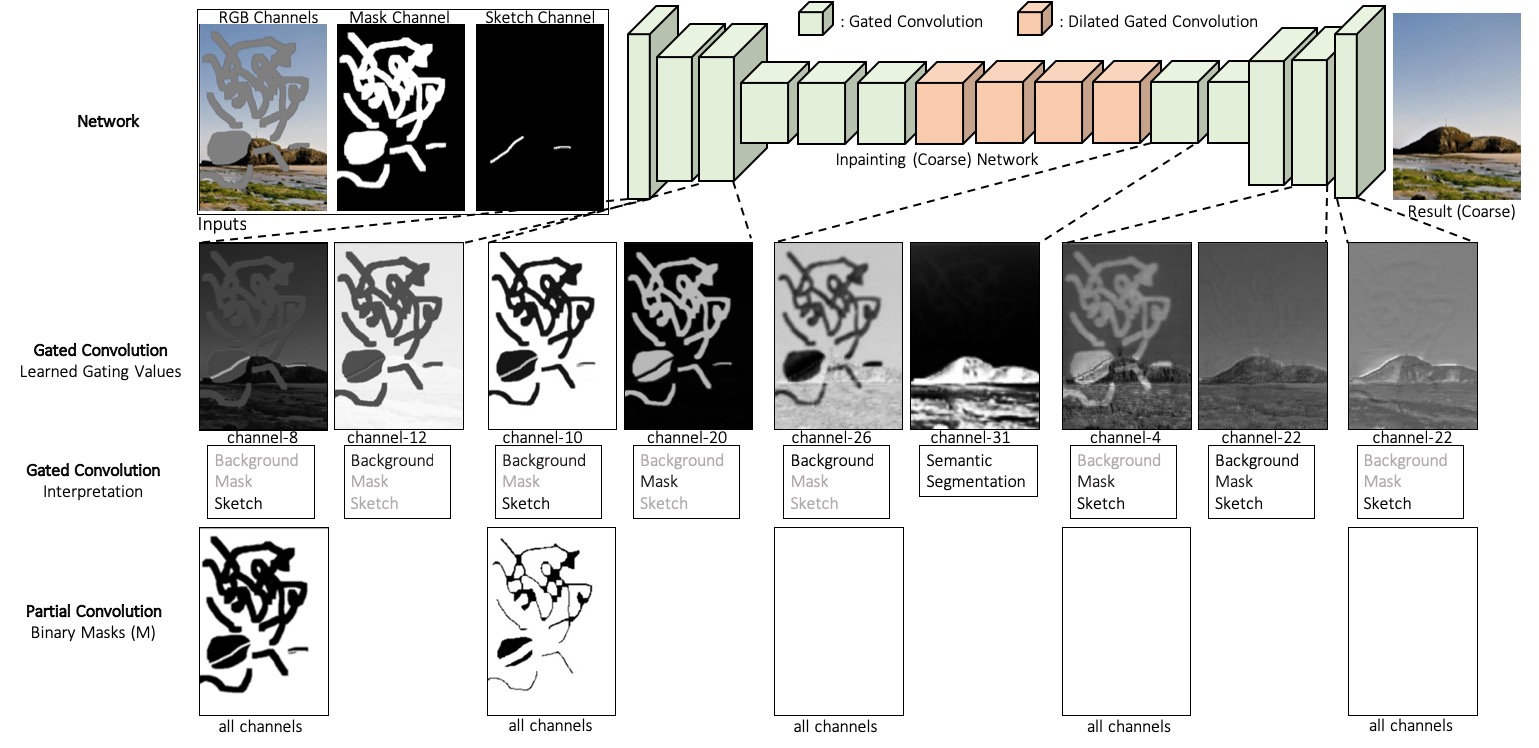}

  \caption{Comparisons of gated convolution and partial convolution with visualization and interpretation of learned gating values. We first show our inpainting network architecture based on~\cite{yu2018generative} by replacing all convolutions with gated convolutions in the 1st row. Note that for simplicity, the following refinement network in~\cite{yu2018generative} is ignored in the figure. With same settings, we train two models based on gated convolution and partial convolution separately. We then directly visualize intermediate un-normalized gating values in the 2nd row. The values differ mainly based on three parts: \textbf{background}, \textbf{mask} and \textbf{sketch}. In the 3rd row, we provide an interpretation based on which part(s) have higher gating values. Interestingly we also find that for some channels (\eg~channel-31 of the layer after dilated convolution), the learned gating values are based on foreground/background semantic segmentation. For comparison, we also visualize the un-learnable fixed binary mask \(M\) of partial convolution in the 4th row.}
  \label{figs:feature_vis}
\end{figure*}
In Figure~\ref{figs:feature_vis}, we provide the visualization and interpretation of learned gating values in our inpainting network, and compare them with that of PartialConv~\cite{liu2018image}.

\section{Ablation Study of SN-PatchGAN} \label{secs:ablation}
\begin{figure*}[ht]
  \centering
  \includegraphics[width=\linewidth]{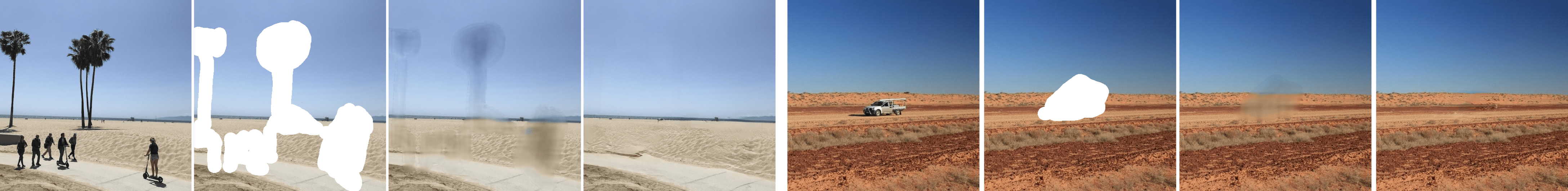}
  \vspace*{-6mm}
  \caption{Ablation Study of SN-PatchGAN. From left to right, we show original image, masked input, results with one global GAN and our results with SN-PatchGAN. SN-PatchGAN is proposed because free-form masks may appear anywhere in images with any shape. Global and local GANs~\cite{iizuka2017globally} designed for a single rectangular mask are not applicable.}
  \label{figs:ablation_study}
\end{figure*}

In this section, we present ablation study to demonstrate the effectiveness of SN-PatchGAN. It is noteworthy that SN-PatchGAN is proposed because free-form masks may appear anywhere in images with any shape. Global and local GANs~\cite{iizuka2017globally} designed for a single rectangular mask are not applicable. Previous work have already shown that (1) one vanilla global discriminator has much worse performance than two local and global discriminators~\cite{iizuka2017globally}, and (2) GAN with spectral normalization has better stability and performance. We also provide experiments of SN-PatchGAN in the context of image inpainting in Figure~\ref{figs:ablation_study}. Our image inpainting network trained on a global GAN without spectral normalization has significantly worse performance on all examples.

\section{More Comparison Results} \label{secs:comparison}
\begin{figure*}[ht]
  \centering
  \vspace*{-9.5mm}
  \includegraphics[width=0.95\linewidth]{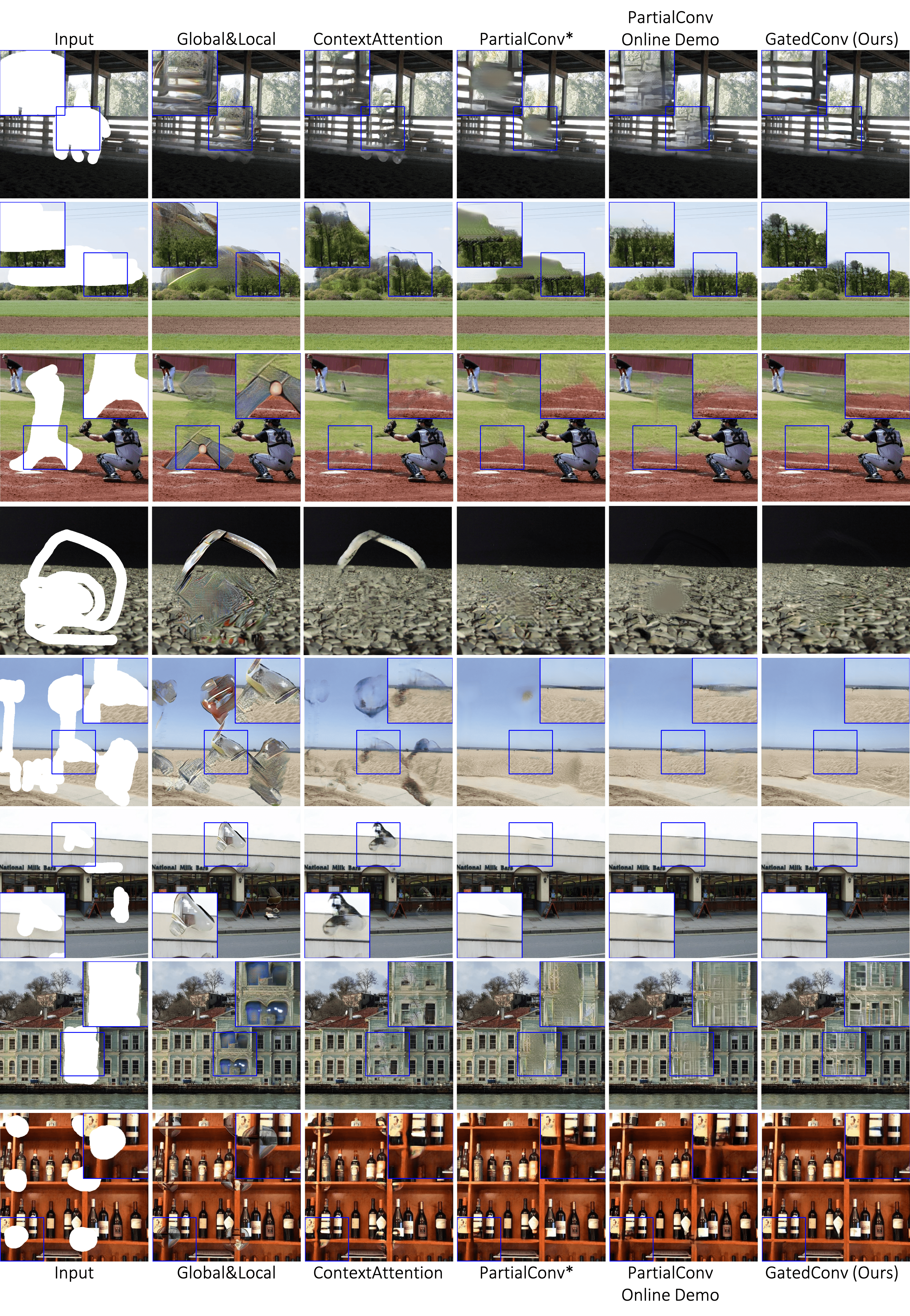}
  \vspace*{-5mm}
  \caption{More comparison results on natural scenes. Best-viewed with zoom-in on PDF to see color shadows and artifacts.}
  \label{figs:supplementary_comparison}
\end{figure*}

\begin{figure*}[ht]
  \centering
  \includegraphics[width=0.9\linewidth]{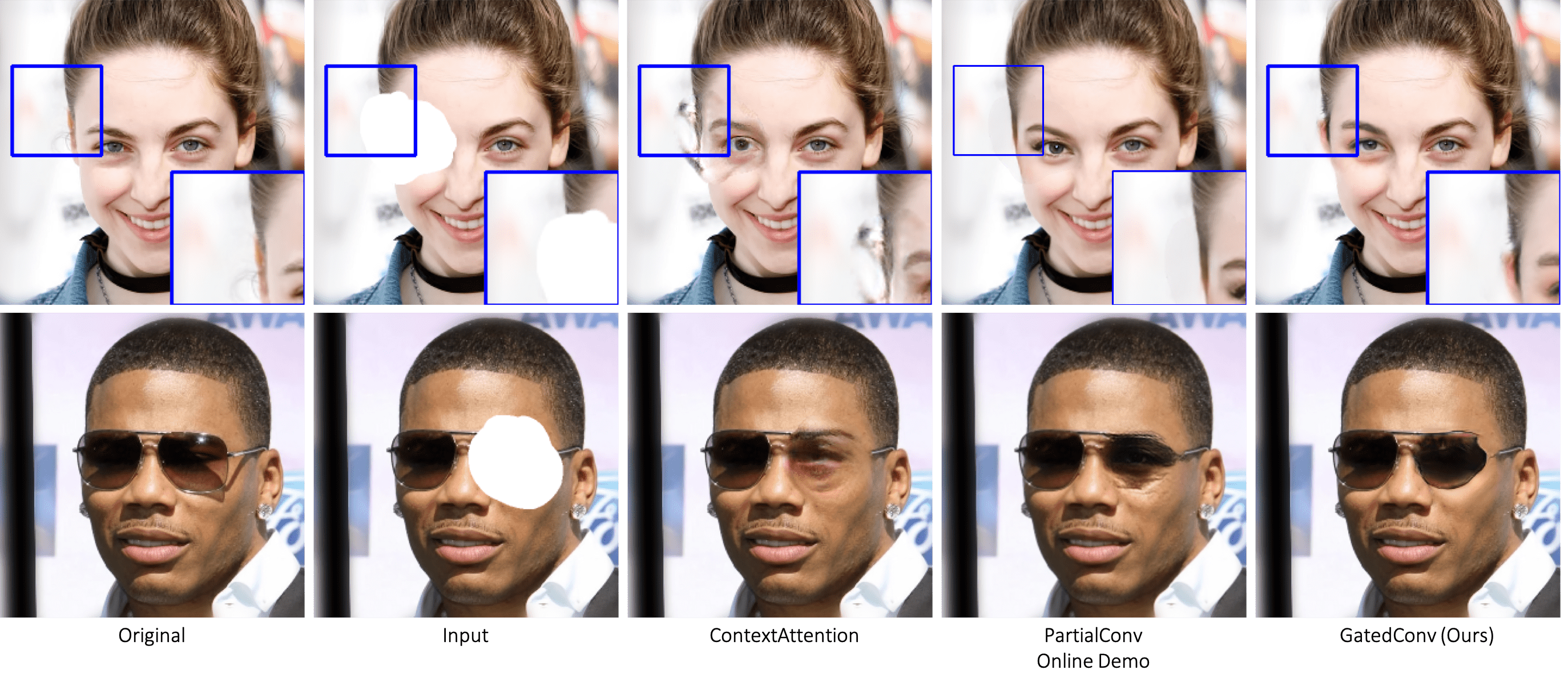}
  \caption{More comparison results on faces. Best-viewed with zoom-in on PDF to see color shadows and artifacts.}
  \label{figs:supplementary_face_comparison}
\end{figure*}

In this section, we show more comparison results of learning-based image inpainting systems including Global\&Local~\cite{iizuka2017globally}, ContextAttention~\cite{yu2018generative}, PartialConv~\cite{liu2018image} (both our implementation within same framework and official model via online demo) and our proposed method based on gated convolution. Note that the models of scenes and faces are trained in separate following all other methods~\cite{iizuka2017globally, liu2018image, yu2018generative}. All testing images are not in the training set. Results are shown in Figure~\ref{figs:supplementary_comparison} and Figure~\ref{figs:supplementary_face_comparison}. Compared with our baseline PartialConv, our inpainting system generates higher-quality inpainting results. Although PartialConv significantly improves over previous baselines like Global\&Local~\cite{iizuka2017globally} and  ContextAttention~\cite{yu2018generative}, it still produces observable color inconsistency or shadows in both official online demo and our reproduced version (best-viewed with zoom-in on PDF to see color shadows and artifacts). Moreover, PartialConv fails especially on cases (1) when holes are large and involving transitions of two segments (\eg, a mask covering both sky and ground), and (2) when the image has strong structure/contour/edge prior. The reasons are discussed in the introduction of main paper that un-learnable rule-based hard-gating heuristically categorizes all input locations to be either invalid or valid, ignoring many other important information. Gated convolution is able to leverage these information by learning a soft-gating end-to-end.

\section{More Inpainting Results of Our System} \label{secs:more_results}
\begin{figure*}[t]
  \centering
  \includegraphics[width=0.9\linewidth]{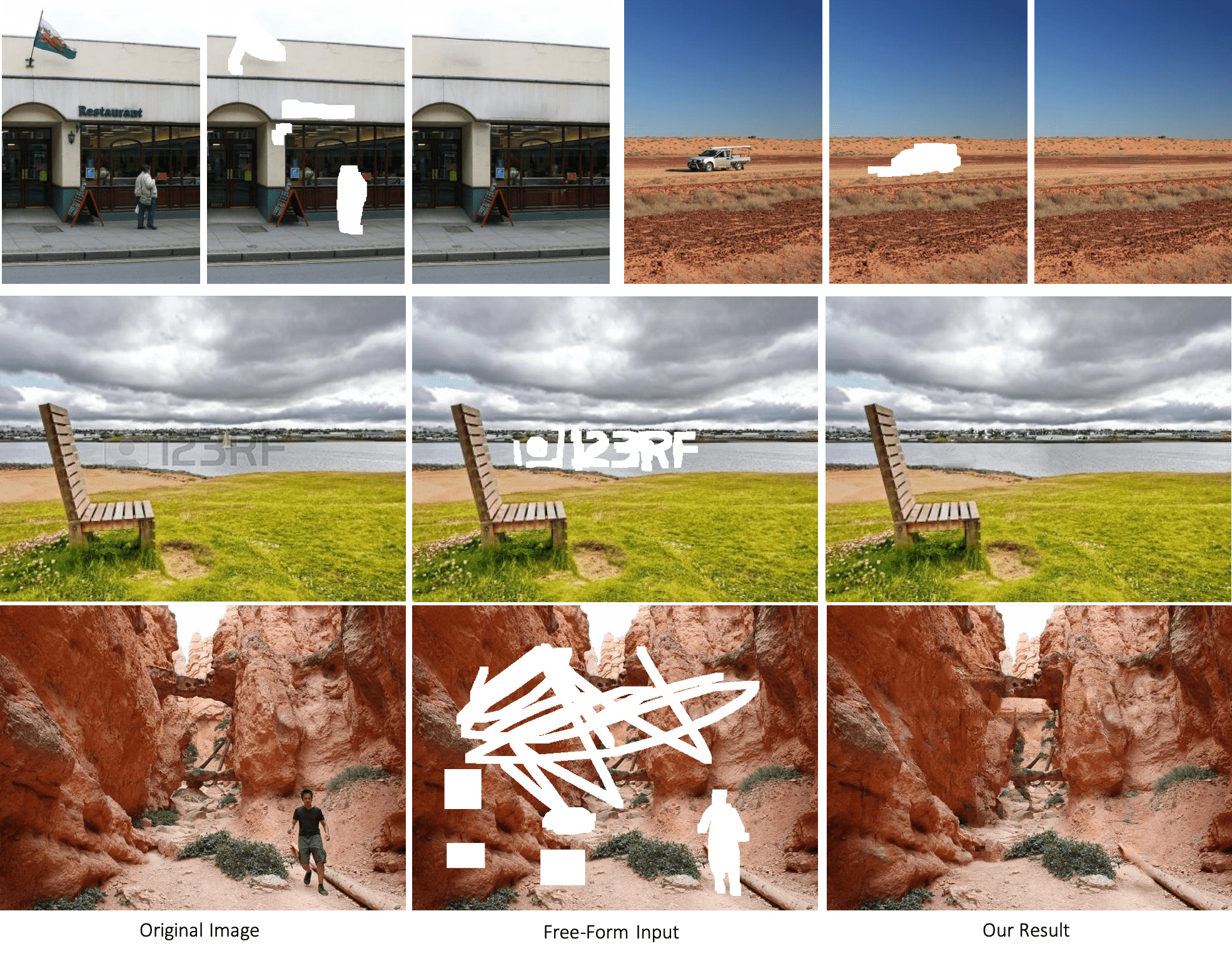}
  \caption{More results from our free-form inpainting system on natural images (1).}
  \label{figs:supplementary_places}
\end{figure*}

\begin{figure*}[t]
  \centering
  \includegraphics[width=\linewidth]{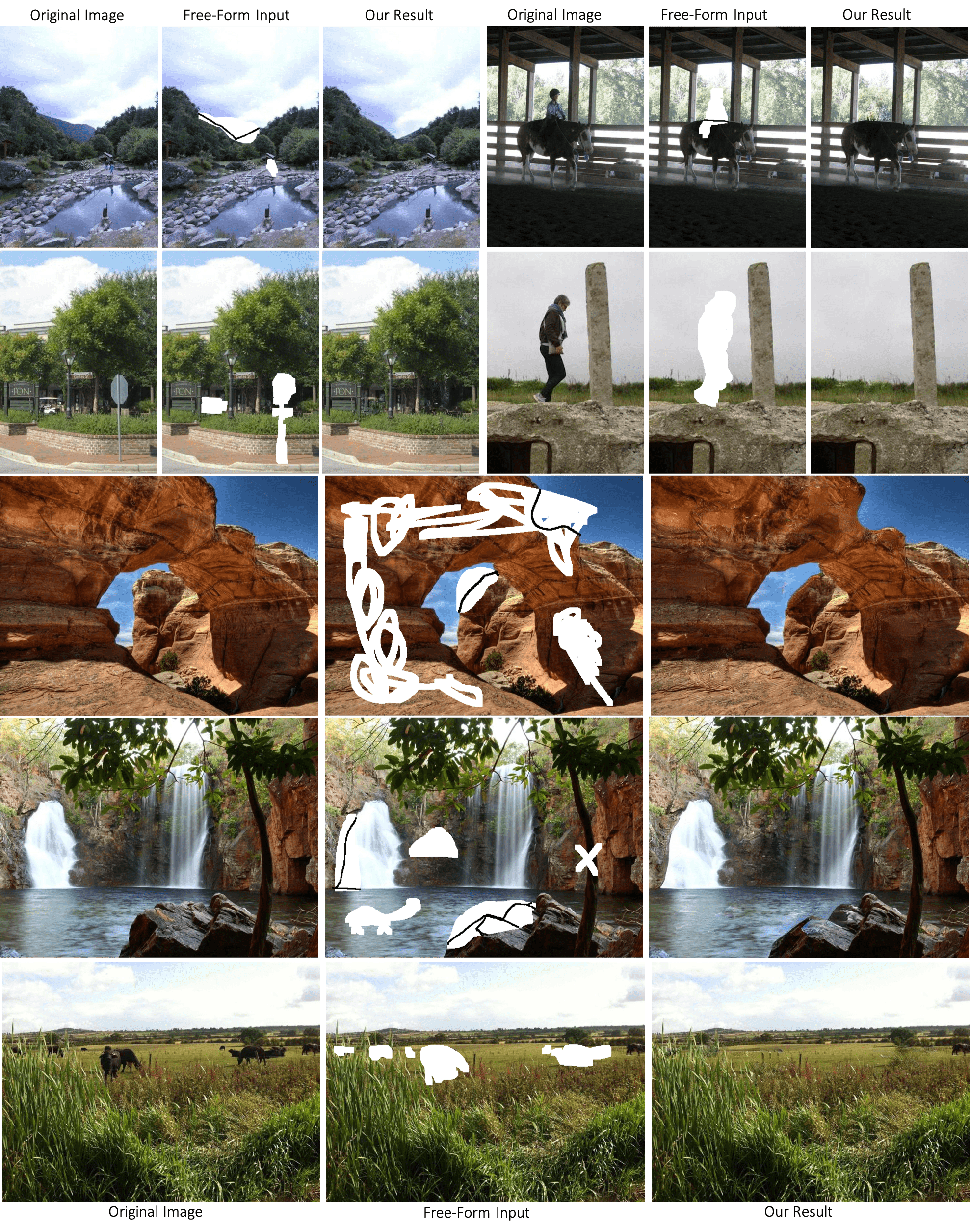}
  \caption{More results from our free-form inpainting system on natural images (2).}
  \label{figs:supplementary_places2}
\end{figure*}

\begin{figure*}[ht]
  \centering
  \includegraphics[width=\linewidth]{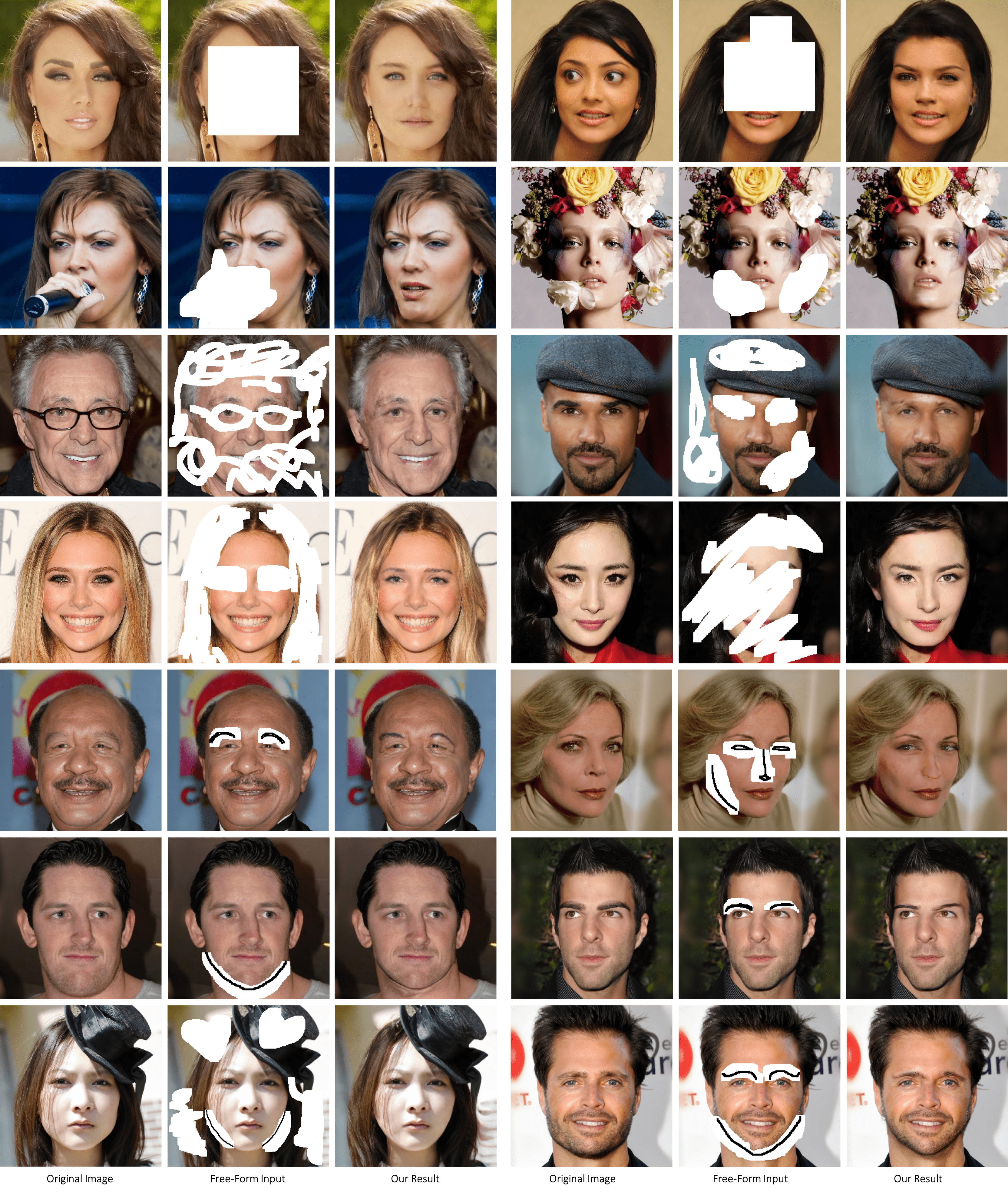}
  \caption{More results from our free-form inpainting system on faces.}
  \label{figs:supplementary_celeba}
\end{figure*}

In this section, we present more examples towards real use cases based on our proposed image inpainting system. We show inpainting results on both natural scenes and faces in Figure~\ref{figs:supplementary_places}, Figure~\ref{figs:supplementary_places2} and Figure~\ref{figs:supplementary_celeba}. We show our inpainting system helps user quickly remove distracting objects, modify image layouts, edit faces and interactively create novel objects in images.

\end{document}